\documentclass[conference]{IEEEtran}
\IEEEoverridecommandlockouts
\usepackage{cite}
\usepackage{amsmath,amssymb,amsfonts}
\usepackage{algorithmic}
\usepackage{graphicx}
\usepackage{textcomp}
\usepackage{xcolor}
\def\BibTeX{{\rm B\kern-.05em{\sc i\kern-.025em b}\kern-.08em
    T\kern-.1667em\lower.7ex\hbox{E}\kern-.125emX}}
    
\usepackage{caption}
\usepackage{subcaption}

\begin{document}

\title{Adversarial Machine Learning Security Problems for 6G: mmWave  Beam Prediction Use-Case}

\author{\IEEEauthorblockN{Evren Catak}
\IEEEauthorblockA{\textit{Norwegian University of}\\ \textit{Science and Technology} \\
Gjøvik, Norway \\
evren.catak@ntnu.no}
\and
\IEEEauthorblockN{Ferhat Ozgur Catak}
\IEEEauthorblockA{\textit{Simula Research Lab.} \\
Fornebu, Norway \\
ozgur@simula.no}
\and
\IEEEauthorblockN{Arild Moldsvor}
\IEEEauthorblockA{\textit{Norwegian University of}\\ \textit{Science and Technology} \\
Gjøvik, Norway \\
arild.moldsvor@ntnu.no}
}

\maketitle

\begin{abstract}
6G is the next generation for the communication systems. In recent years, machine learning algorithms have been applied widely in various fields such as health, transportation, and the autonomous car. The predictive algorithms will be used in 6G problems. With the rapid developments of deep learning techniques, it is critical to take the security concern into account to apply the algorithms. While machine learning offers significant advantages for 6G, AI models' security is ignored. Since it has many applications in the real world, security is a vital part of the algorithms. This paper has proposed a mitigation method for adversarial attacks against proposed 6G machine learning models for the millimeter-wave (mmWave) beam prediction with adversarial learning. The main idea behind adversarial attacks against machine learning models is to produce faulty results by manipulating trained deep learning models for 6G applications for mmWave beam prediction use case. We have also presented the adversarial learning mitigation method's performance for 6G security in millimeter-wave beam prediction application with fast gradient sign method attack. The mean square errors of the defended model and undefended model are very close.

\end{abstract}


\section{Introduction}
In the past 20 years, most of the physical layer technologies, i.e., modulations, multiple access waveforms, coding techniques and time/frequency multiplexing, have flourished over the evolution of cellular systems. However, up to 4G, time-frequency domain technology has been explored to increase overall system capacity \cite{9040431}, \cite{9321326}. The recent developments in 5G and beyond technologies support emerging applications such as smart homes, vehicular networks, augmented reality (AR), virtual reality (VR) with unprecedented rates enabled by recent advances in massive multiple-input multiple-output (MIMO), millimeter-wave (mmWave) communications, network slicing, small cells, and Internet of things (IoT). These complex structures of 5G and beyond technologies can be captured by using data-driven approach machine learning (ML) algorithm \cite{7792374,8666641, 8972389}. The strong learning, reasoning, intelligent recognition abilities of ML allow the network structure to train and adapt itself to support the diverse demands of the systems without human intervention \cite{8360430}.

The extraordinary growth of data traffic on wireless communication has driven the need to examine the highest frequency spectrum to meet the requirements by using mmWave communications \cite{6736750}. The frequency range of the mmWave communication system is between 30 and 300 GHz the available bandwidth is about 250 GHz. Enabling mmWave communication faces mainly three critical challenges \cite{8395149, 7400949} i) the sensitivity for atmospheric attenuation obligates it to propagate solely by line-of-sight paths, ii) hand over problem between base stations (BSs), iii) adjustment of the large numbers of beamforming arrays. In addition, due to the use of large antenna arrays and low complexity, transceiver demands are captured by using ML algorithms for mmWave communication.

mmWave communication systems require the pointing of the narrow beams. The goal is to choose the best beams for the analogue beamforming with both receiver and transmitter having multi-antenna arrays. A beam codeword is a set of analogue phase-shifted values applied to the antenna elements due to an analogue beam \cite{8767936}\cite{9048614}. In \cite{9034044}, deep learning base beam selection is proposed that exploits channel state information for the sub-6 GHz links. In addition to beam prediction, information about the locations and sizes of vehicles in the communication environment are used in \cite{8644288} to predict the optimal beam pair. Locational based beamforming solutions are more suitable for line-of-sight (LOS) communication. The same locations for the non-line-of-sight (NLOS)transmission need different beamforming solution.

The integration of the ML for the 6G beyond technologies lead to potential security concerns \cite{9229146}, \cite{sagduyu2021adversarial}. Especially, wireless communication systems have security vulnerabilities due to their nature. The studies 6G beyond technologies with ML methods should be evaluated in terms of security. Current research is just building the AI models for the 6G communication problems. On the other hand, security concerns are ignored in previous studies. Based on the shortcomings of the literature's security concepts, we deal with the security problem of machine learning application for beamforming prediction. Alkhateeb et al. \cite{8395149} proposed a feed-forward deep learning model for RF beamforming codeword prediction with several base stations (BSs) with multiple users. The BSs beamforming vectors are predicted from the received signals using the omni and quasi-omni beam patterns to enable both LOS and NLOS transmissions. While the proposed method in \cite{8395149} showed promising results for the beamforming problem, the deep learning algorithm itself security was not investigated. Based on ML-based beamforming prediction's shortcomings, in this study, we focus on adversarial attack strategies based on loss maximisation-based attacks against proposed AI models for 6G mmWave communication. We consider the adversarial machine learning attacks to poisoning the beamforming prediction model \cite{8395149}. Thus our main contributions for this paper are as follows:
\begin{itemize}
	\item We show that an undefended RF beamforming codeword deep learning model's prediction performance will decrease with the craftily designed adversarial noise.
	\item We demonstrated that the adversarial training based robustness approach is one of the mitigation methods for this domain.
\end{itemize}
The rest of the paper is organized as follows: Section \ref{sec:preliminaries} describes preliminary information about model uncertainty and uncertainty quantification. Section \ref{sec:system_overview} shows our system overview. Section \ref{sec:experiments} evaluates the proposed uncertainty quantification method. Section \ref{sec:conclusion} concludes this paper.

\vspace{2pt}

\emph{Notations}

In this paper, we employ the following notations:
\begin{itemize}
\item Vectors denote in lowercase bold font and matrices in uppercase bold font i.e., $ \textbf{a} $ and $ \textbf{A} $ respectively.  
\item For a given vector $ \textbf{a} $, $ \textbf{a}_{i,j} $ and $ \textbf{a}_{k} $ denote $ (i,j) $-th component of $ \textbf{a} $ and $ k $-th component of $ \textbf{a} $ respectively. For a integer $ d $, $ d_{i,j} $ means $ (i,j) $-th components of matrix $ \textbf{D} $.
\end{itemize}
\section{Preliminary Information} \label{sec:preliminaries}

\subsection{Downlink Transmission}
\begin{figure}[t!]
    \centering
    \includegraphics[width=1.0\linewidth]{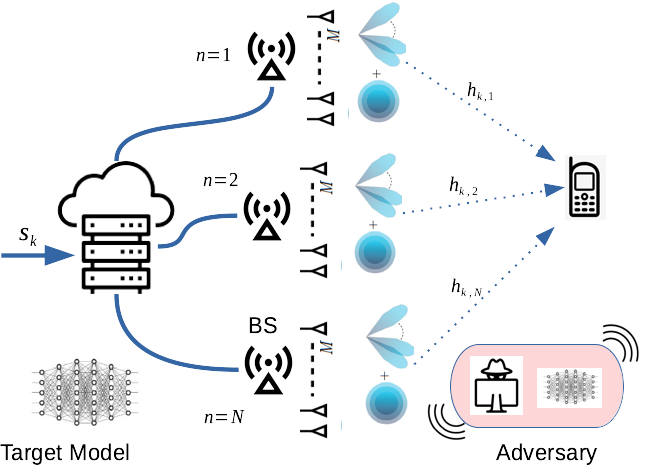}
    \caption{Block diagram of the mmWave beamforming system.}
    \label{fig:downlink_comm}
\end{figure}

Let consider a mmWave communication system as in Figure \ref{fig:downlink_comm} where $N$ is the number of the BSs with equipped $M$ antennas are serving for one mobile user who has single antenna. All the BSs are connected with a cloud processing unit. The transmitted signal $  \textbf{s} = [s_1, s_2,\dots, s_K]$ with $K$ subcarriers is firstly precoded by using code vector $  \textbf{c}_{k}= [c_{k,1}, c_{k,2},\dots, c_{k,N}]^T$ and then is transformed into time domain with using $K$-point IFFT operation. Thus, the baseband signal from the $n$-th BS and $k$-th subcarrier is 
\begin{equation}\label{Eqn:b1}
	\textbf{x}_{k,n} = \textbf{f}_{n}c_{k,n}s_k 
\end{equation}
where $\textbf{f}_{n}$ is the beam steering vector is defined for each BS antennas as [$\textbf{f}_{n}]_m =  \frac{1}{\sqrt{M}}e^{j\theta_{n,m}}$ where $\theta_{n,m}$ is a quantized angle. RF precoding matrix $\textbf{F}^RF= blkdiag(\textbf{c}_{1}, \textbf{c}_{2\dots,\textbf{c}_{N}}) \in \mathbf{C}^{NM\times N}$ The received signal at the $k$-th subcarrier is expressed as
\begin{equation}\label{Eqn:b2}
	\textbf{y}_{k} =\sum_{n=1}^{N} \textbf{h}_{k,n}^T \textbf{x}_{k,n} + v_k 
\end{equation}
where $v_k$ is additive white Gaussian noise (AWGN) with variance $ \sigma ^{2} $, i.e., $N(0, \sigma^2)$.
\subsection{Effective Achievable Rate}
Perfect channel information satisfies optimum achievable rate, however, the channel state information requires large training overhead due to the large number of the antennas. On the other hand, the channel information and beamforming vector need to be updated as the user moves. This issues can be captured as with the channel coherence time $T_{C}$, and channel beam coherence time $T_{B}$, which are examined in detail in \cite{7742901}. The multi-path channel and beams stay aligned on the $T_{C}$ and $T_{B}$ duration respectively. The channel training and beamforming design take place in the first $T_{tr}$, the rest of it is used to the data transmission. To develop a model with efficient channel training and beamforming design, the effective achievable rate need to be maximized. The final problem formulations \cite{8395149} are 
\begin{equation}\label{Eqn:b3}
\begin{aligned}
& \prod \left(T_{tr},\lbrace \textbf{c}_{k}\rbrace_{k=1}^{K},\textbf{F}^R, \mathcal{F} \right)=  \\ 
    & {\mathrm{argmax}} \left( 1- \frac{T_{TR}}{T_{B}} \right)  \sum_{k=1}^{K}  \log_2  \left(  1+ SNR\vert  \sum_{n=1}^{N}   \textbf{h}_{k,n}^T \textbf{f}_{n}c_{k,n} \vert^2  \right)
   \end{aligned}
\end{equation}

\begin{equation}\label{Eqn:b4}
s.t.\, \, \, \textbf{f}_{n} \in \mathcal{F}, \, \, \forall n 
\end{equation}
\begin{equation}\label{Eqn:b5}
\| c_{k}\| ^2 = 1 \ \ \forall k
\end{equation}
where $\mathcal{F}$ is the quantized codebook for the BSs RF beamforming vectors. Solving these equations determine a solution for a low channel training ahead and realize the beamforming vector to satisfy the maximum achievable rate, $R$. 

\subsection{Using Deep Learning Algorithms to estimate RF beamforming vectors}
Using the benefits of machine learning algorithms is a novel solution for a massive amount of MIMO channel training and scanning a large number of narrow beams. The beams depend on the environmental conditions like user and BSs locations, furniture, trees, building e.t.c. It is too difficult to define these environment conditions as a closed-form equation. The solution is to use omni and quasi-omni beam pattern to predict the best RF beamforming vectors. Using these beam patterns benefits to take into account the reflection and diffraction of the pilot signal.

The deep learning solution consists of two states: training and prediction. Firstly, the deep learning model learns the beams according to the omni-received pilots. Secondly, the model uses the trained data to predict the RF beamforming vector for the current condition. 
\subsubsection{Training Steps}The user sends uplink training pilot sequences for each beam coherence time $T_{B}$. BSs combine received pilot sequences on RF beamforming vector $B$ and fed back to the cloud. The cloud uses the received sequences from all the BSs as the input of the deep learning algorithm to find the achievable rate in (\ref{Eqn:b6}) for every RF beamforming vector to represent the desired outputs, i.e. $\textbf{g}_{p}$ is the channel coefficient from ommi beam.
\begin{equation}\label{Eqn:b6}
	{R}_{n} ^{(p)}=\frac{1}{K} \sum_{n=1}^{N}  \log_2  \left(  1+ SNR\vert \textbf{h}_{k,n}^T \textbf{g}_{p} \vert^2  \right)
\end{equation}
\subsubsection{Learning Steps} In this stage, the trained deep learning model is used to prediction the RF beamforming vectors. Firstly, the user sends an uplink pilot sequence. The BSs combine these sequences and send them to the cloud. Then, the cloud uses the trained deep learning model to predict the best RF beamforming vectors to maximize the achievable rate for each BS. Finally, BSs use the predicted RF beamforming vector to estimate the effective channel $\textbf{h}_{k,n}$.

To sum up, machine learning algorithms find diverse applications in a wireless communication system that we consider the RF beamforming vector prediction \cite{8395149}. On the other hand, security concerns in wireless communication are also a problem for the ML algorithm.
In the following section, we will briefly describe adversarial machine learning, attack environments, and adversarial training that we have used in this study.
\subsection{Adversarial Machine Learning}
Adversarial machine learning is an attack technique that attempts to fool neural network models by supplying craftily manipulated input with a small difference. The obvious intention is to produce a failure in a neural network model\cite{2016arXiv161101236K}.
Attackers apply model evasion attacks for phishing attacks, spams, and executing malware code in an analysis environment \cite{8965459}. There are also some advantages to attackers in misclassification and misdirection of models. In such attacks, the attacker does not change training instances. Instead, he tries to make some small perturbations in input instances in the model's inference time to make this new input instance seem safe (i.e. normal behaviour) \cite{2021arXiv210204150F}. We mainly concentrate on this kind of adversarial attacks in this study. There are many attacking methods for deep learning models, and the Fast-Gradient Sign Method (FGSM) is the most straightforward and powerful attack type. We only focus on the FGSM attack, but our solution to prevent this attack can be applied to other adversarial machine learning attacks.
%%% My Previous works

\begin{figure}[h]
    \centering
    \includegraphics[width=1.0\linewidth]{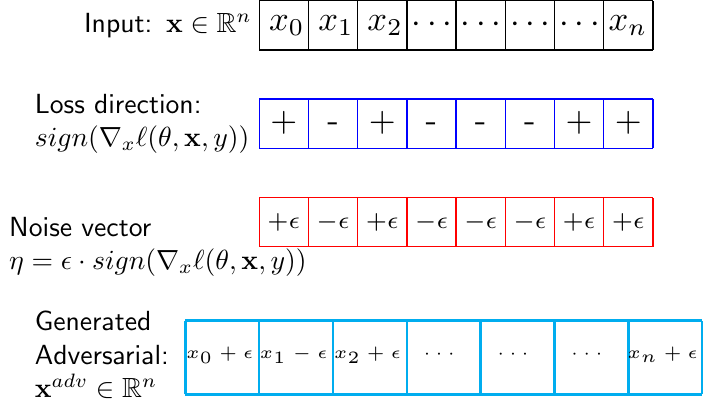}
    \caption{FGSM attack steps. The input vector $\mathbf{x} \in \mathbb{R}^n$ is poisoned with loss maximization direction. }
    \label{fig:fgsm_detail}
\end{figure}

FGSM works by utilizing the gradients of the neural network to create an adversarial example to evade the model. For an input instance $\mathbf{x}$, the FGSM utilizes the gradients $\nabla_x$ of the loss value $\ell$ for the input instance to build a new instance $\mathbf{x}^{adv}$ that maximizes the loss value of the classifier hypothesis $h$. This new instance is named the adversarial instance. We can summarize the FGSM using the following explanation:
\begin{equation}
	\mathbf{x}^{adv} = \mathbf{x} + \epsilon \cdot sign(\nabla_x \ell(\mathbf{\theta},\mathbf{x},y))
\end{equation}
By adding a slowly modest noise vector $\eta \in \mathbb{R}^n$ whose elements are equal to the sign of the features of the gradient of the cost function $\ell$ for the input $\mathbf{x} \in \mathbb{R}^n$, the attacker can easily manipulate the output of a deep learning model.
Figure \ref{fig:fgsm_detail} shows the details of the FGSM attack.
\subsection{Adversarial Training}
Adversarial training is a widely recommended defense that implies generating adversarial instances using the gradient of the victim classifier, and then re-training the model with the adversarial instances and their respective labels. This technique has demonstrated to be efficient in defending models from adversarial attacks. 

Let us first think a common classification problem with a training instances $X \in \mathbb{R}^{m \times n}$ of dimension $d$, a label space $Y$ We assume the classifier $h_\theta$ has been trained to minimize a loss function $\ell$as follows:
\begin{equation}
	\label{eq:cost-func}
	\underset{\theta}{min}\frac{1}{m} \sum_{i=1}^{m} \ell(h_\theta(\mathbf{x}_i,y_i))
\end{equation}
Given a classifier model $h_\theta(\cdot)$ and an input instance $x$, whose responding output is $y$, an adversarial instance $x^*$ is an input such that:
\begin{equation}
	\label{eq:adv_ex}
	h_\theta(x^*) \neq y \,\,\,\,\, \wedge \, d(x,x^*) < \epsilon
\end{equation}
where $d(\cdot,\cdot)$ is the distance metric between two input instances original input $x$ and adversarial version $x^*$. Most actual adversarial model attacks transform Equation \ref{eq:adv_ex} into the following optimization problem:
\begin{equation}
	\underset{x}{\mathrm{\textbf{argmax}}} \, \ell \left(h_\theta(x^*),y\right)
\end{equation}
\begin{equation}
	s.t. d(x,x^*) < \epsilon
\end{equation}
where $\ell$ is loss function between predicted output $h(\cdot)$ and correct label $y$.

In order to mitigate such attacks, at per training step, the conventional training procedure from Equation \ref{eq:cost-func} is replaced with a \texttt{min-max} objective function to minimize the expected value of the maximum loss, as follows:
\begin{equation}
	\underset{\theta}{min} \, \underset{(x,y)}{\mathbb{E}} \left(\underset{d(x,x^*)<\epsilon}{max} \ell(h(x^*),y) \right)
\end{equation}

\section{System Model}\label{sec:system_overview}
\subsection{Adversarial Training}
\begin{figure}[htbp!]
    \centering
    \includegraphics[width=1.0\linewidth]{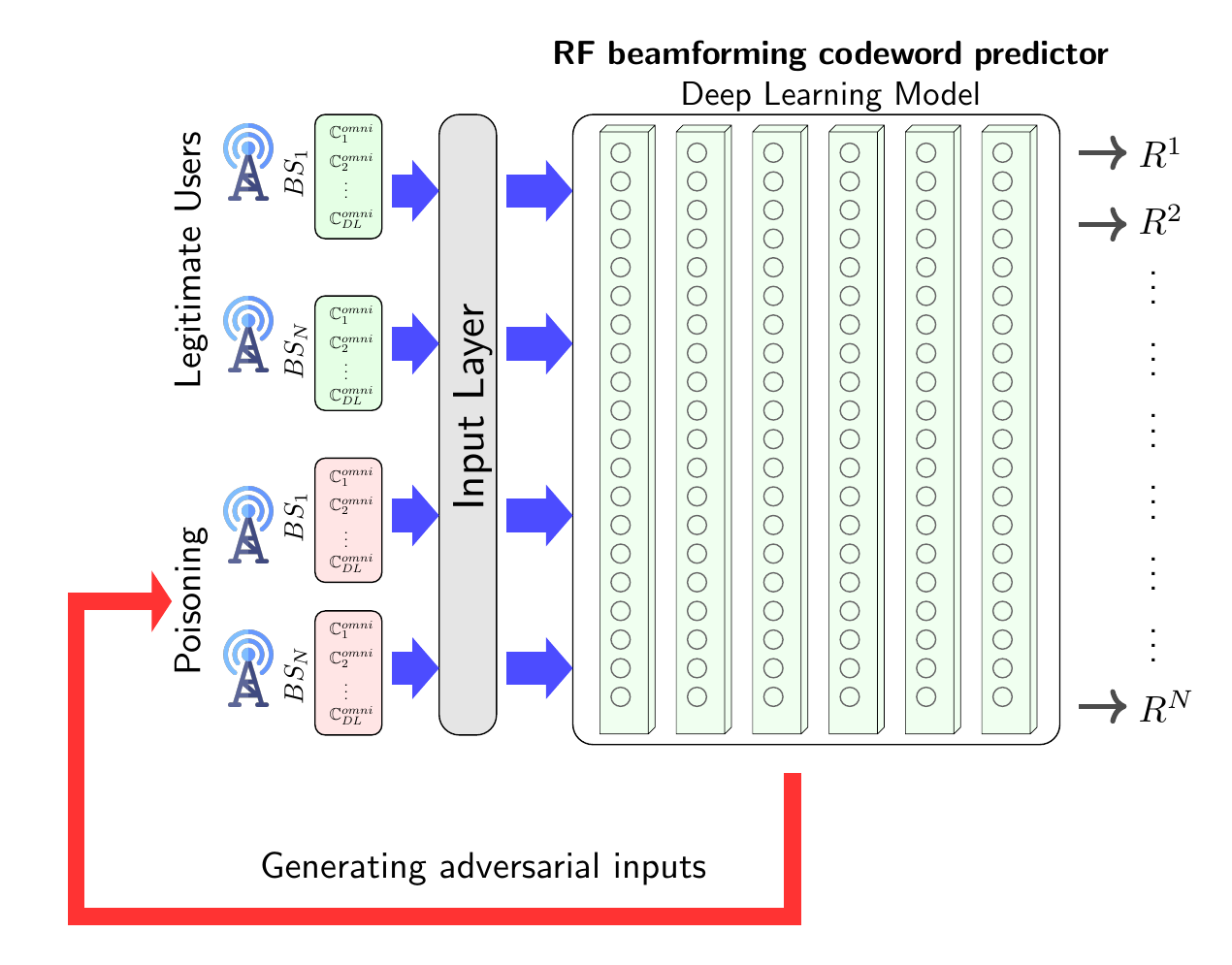}
    \caption{The diagram of RF beamforming codeword adversarial training.}
    \label{fig:deepmimo_adv_learning}
\end{figure}

Figure \ref{fig:deepmimo_adv_learning} shows the adversarial training process. After the model is trained, adversarial inputs are created using the model itself, combined with legitimate users information and added to the training. When the model reaches the steady-state state, the training process is completed. In this way, the model will both predict RF beamforming codeword for legitimate users while at the same time being immune to the craftily designed noise attack that will be added as input.
\subsection{Capability of the Attacker}
We assumed that the attacker's primary purpose is to manipulate the RF model by applying a carefully crafted noise to the input data.  In a real-world scenario, this white-box setting is the most desired choice for an attacker that does not take the risks of being caught in a trap. The problem is that it requires the attacker to access the model from outside to generate adversarial examples. After manipulating the input data, the attacker can exploit the RF beamforming codeword prediction model's vulnerabilities in the same manner as in an adversary's sandbox environment. The prediction model predicts the adversarial instances when the attacker can convert some model's outputs as other outputs (i.e. wrong prediction).

However, to prevent this noise addition from being easily noticed, the attacker must answer an optimization problem to determine which regions in the input data (i.e. beamforming) must be modified. By solving this optimization problem using one of the available attack methods  \cite{8965459}, the attacker aims to reduce the prediction performance on the manipulated data as much as possible. In this study, to limit the maximum allowed perturbation allowed for the attacker, we used $l_\infty$ norm, which is the maximum difference limit between original and adversarial instance.

\begin{figure}[htbp!]
\centering
\includegraphics[width=1.0\linewidth]{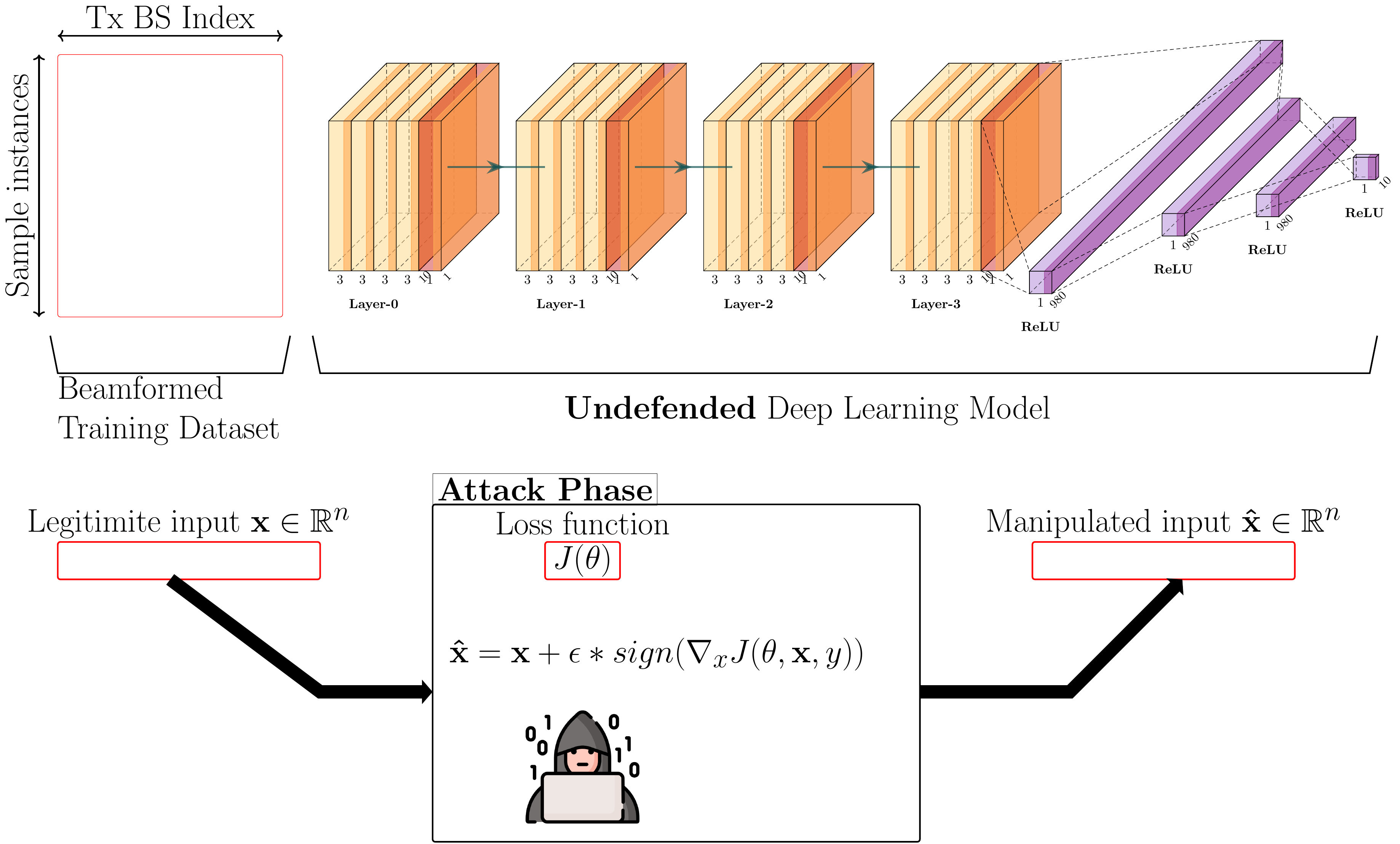}
\caption{Original milimeter-wave beam prediction model results}
\label{fig:adv-ml}
\end{figure}

Figure \ref{fig:adv-ml} shows the attack scenario. The attacker gets an legitimate input, $\mathbf{x}$, creates a noise vector with an $\epsilon$ budget $\eta = \epsilon \cdot sign(\nabla_x \ell(\mathbf{\theta},\mathbf{x},y))$, sums the input instance and the craftily designed noise to create adversarial input $\mathbf{x}^{adv} = \mathbf{x} + \eta$.
\section{Experiments}\label{sec:experiments}
In the experiments, we tested three different scenarios
\begin{itemize}
    \item \textbf{SC1:} Undefended beamforming codework prediction model without any adversary
    \item \textbf{SC2:} Undefended beamforming codework prediction model with FGSM attack
    \item \textbf{SC3:} Adversarial trained beamforming codework prediction model with FGSM attack
\end{itemize}
The experiments were performed using the Python scripts and machine learning libraries: Keras, Tensorflow, and Scikit-learn, on the following machine: 2.8 GHz Quad-Core Intel Core i7 with 16GB of RAM. For all scenarios, two models, undefended and adversarial trained, were built to obtain prediction results. In the first model, the model is trained without any input poisoning. The first model (i.e. undefended model) was used with legitimate users (for SC1) and adversaries (for SC2). The second model (i.e. the adversarially trained model) was used under the FGSM attack. The hyper-parameters such as the number of hidden layers and the number of neurons in the hidden layers, the activation function, the loss function, and the optimization method are the same for both models.

The model architectures are given in Table \ref{tab:dl_arch} and the hyper-parameters selected in Table \ref{tab:dl_arch_params}.
\begin{table}[htbp!]
\centering
\caption{Model architecture}
\label{tab:dl_arch}
\begin{tabular}{c|c}
	\hline \textbf{Layer type} & \textbf{Layer information} \\ \hline \hline
	Fully Connected + ReLU & 100 \\
	Fully Connected + ReLU & 100 \\
	Fully Connected + ReLU & 100 \\
	Fully Connected + TanH & 1 \\ \hline
\end{tabular}
%%%%%%%%%%%%
\end{table}
\begin{table}[htbp!]
\centering
\caption{Milimater-wave beam prediction model parameters}
\label{tab:dl_arch_params}
\begin{tabular}{c|c} 
	\hline
	\textbf{Parameter} & \textbf{Value} \\
	\hline \hline
	Optimizer & Adam \\
	Learning rate & 0.01 \\
	Batch Size & 100 \\
	Dropout Ratio & 0.25 \\
	Epochs & 10 \\
	\hline
\end{tabular}
\end{table}
%%%%%%%%%%%%
\subsection{Research Questions}\label{RQ}
We consider the following two research questions (RQs):
\begin{itemize}
    \item \textbf{RQ1}: Is the deep learning based RF beamforming codeword predictor vulnerable for adversarial machine learning attacks?
    \item \textbf{RQ2}: Is the iterative adversarial training approach a mitigation method for the adversarial attacks in beamforming prediction?
\end{itemize}
\subsection{RF Beamforming Data Generator}\label{sec:datagenerator}
\begin{figure}[htbp!]
    \centering
    \includegraphics[width=1.0\linewidth]{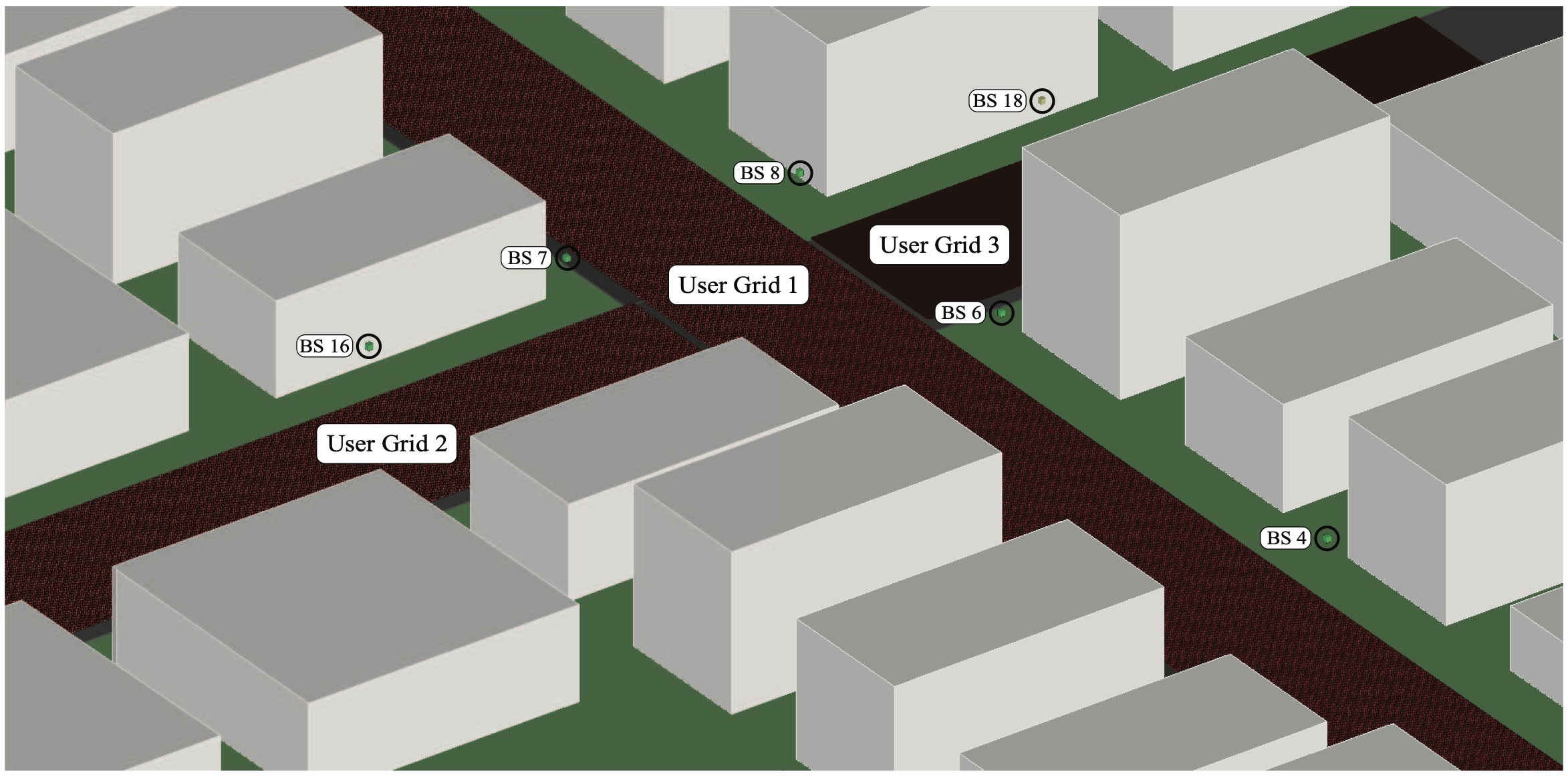}
    \caption{Original scenario \cite{alkhateeb2019deepmimo}. }
    \label{fig:scenario}
\end{figure}

We employed the generic deep learning dataset for millimeter-wave and massive MIMO applications (DeepMIMO) data generator in our experiments \cite{alkhateeb2019deepmimo}. Figure \ref{fig:scenario} shows the bird's-eye view of a section of the O1' ray-tracing scenario, showing the two streets' intersection.
In this section, we conduct experiments on the mmWave communication and massive MIMO applications dataset from the publicly available data set repository. We implemented the proposed mitigation method using Keras and TensorFlow libraries in the Python environment. 
\begin{table}[htbp!]
\centering
\caption{Adversarial settings of our experiments perturbation budget $\epsilon$}
\label{tab:adversarial_settings}
\begin{tabular}{c|c|c} 
	\hline
	\textbf{Attack} & \textbf{Parameters} & \textbf{$l_p$ norm}\\
	\hline \hline
	FGSM & $\epsilon \in [0.01, \cdots 0.1]$ & $l_\infty$\\
	\hline
\end{tabular}
\end{table}

\subsection{Results for RQ1}\label{sec:res_rq1}

\begin{figure}[h]
    \centering
    \includegraphics[width=0.8\linewidth]{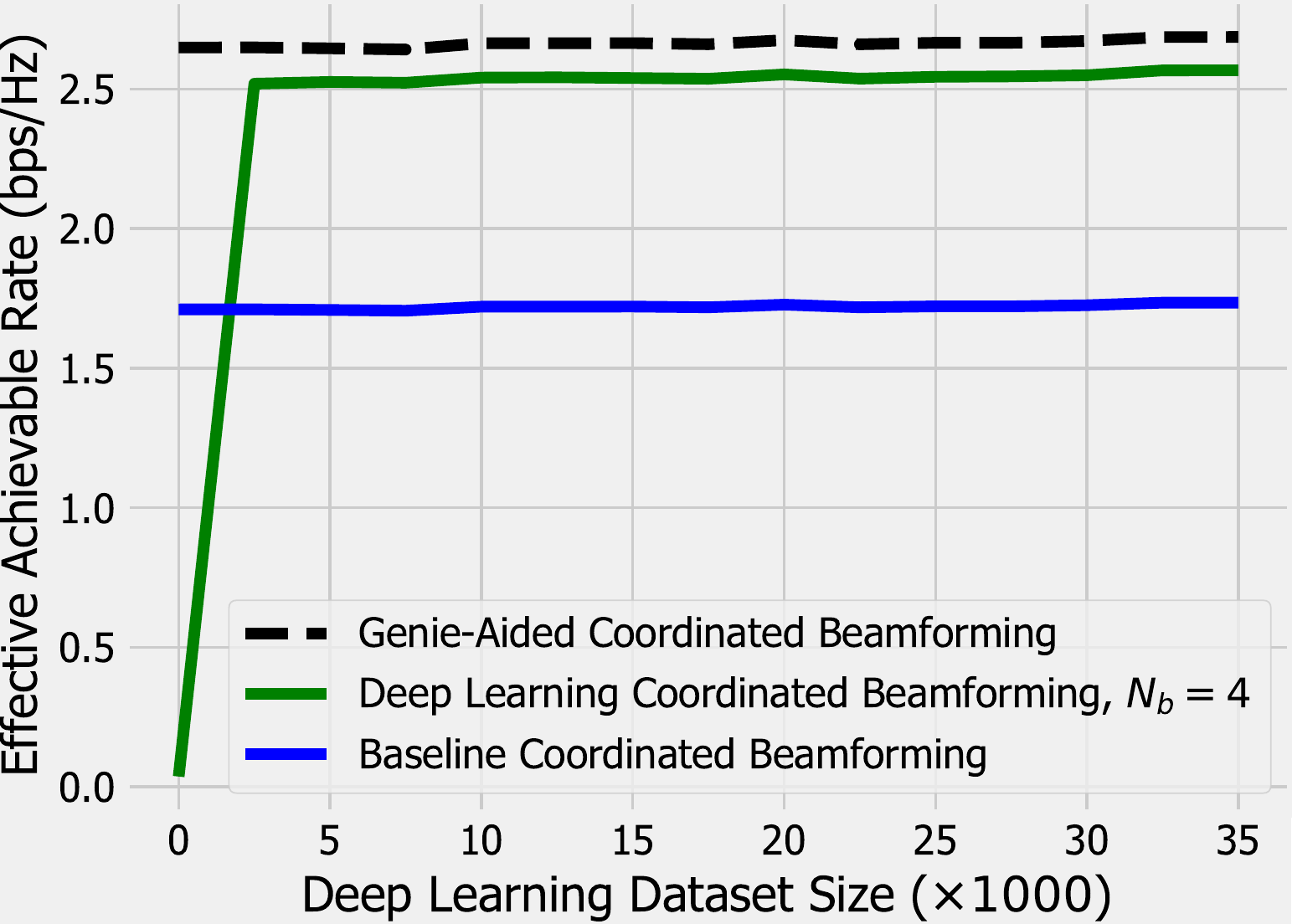}
    \caption{Original (Undefended) RF beamforming codeword deep learning model results. }
    \label{fig:org}
\end{figure}

Figure \ref{fig:org} shows the original undefended deep learning model results without any attack. According to the figure, the deep learning model's predictions are very close the original value.

\begin{figure}[h]
    \centering
    \includegraphics[width=0.8\linewidth]{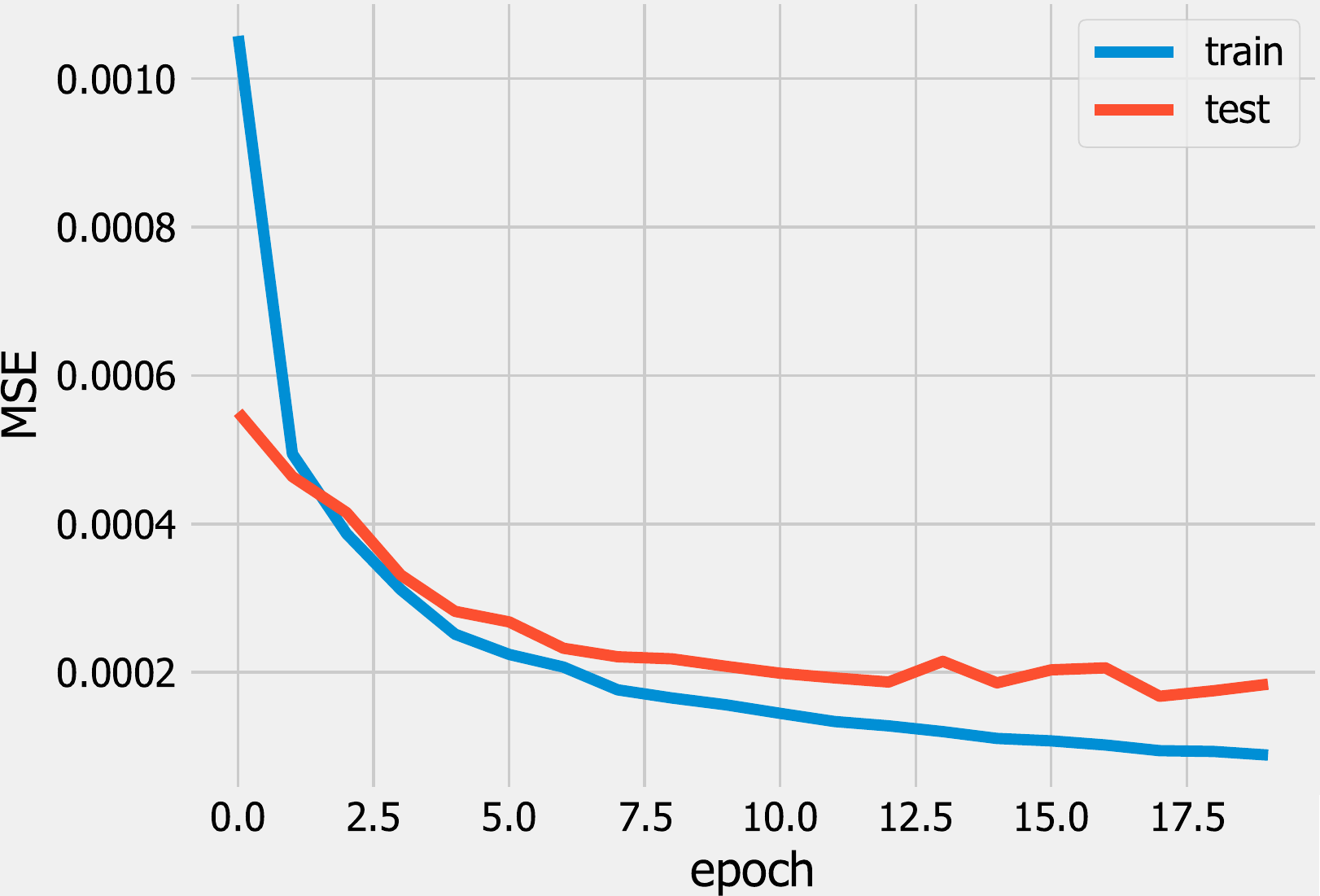}
    \caption{The beamforming prediction model history.}
    \label{fig:history_org}
\end{figure}

Figure \ref{fig:history_org} shows the training history of the beamforming prediction model with 35.000 training instances. The model is trained with clean (non-perturbated) instances.

\begin{figure}[h]
    \centering
    \includegraphics[width=0.8\linewidth]{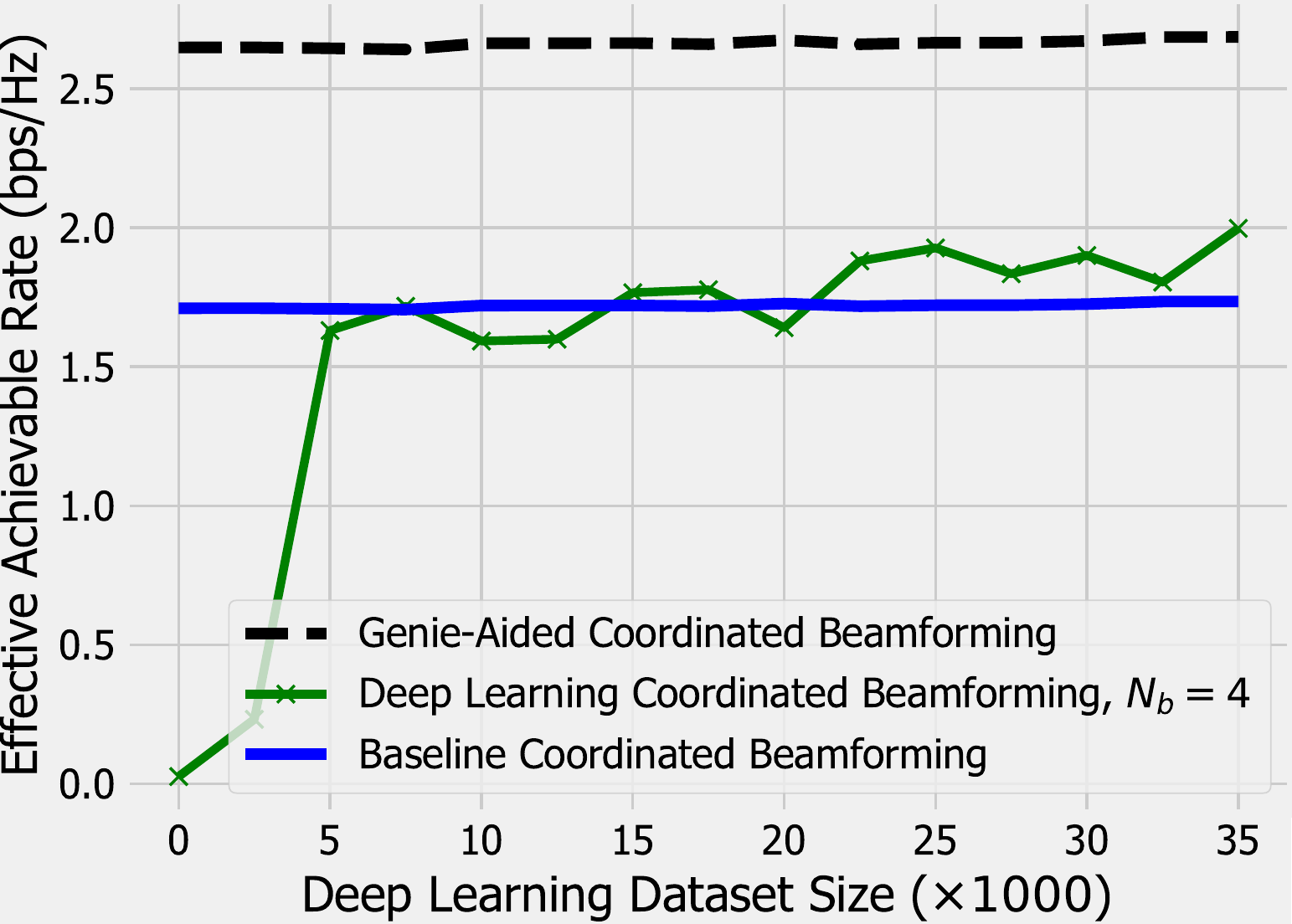}
    \caption{Attacked (Undefended) RF beamforming codeword deep learning model results. }
    \label{fig:attacked}
\end{figure}

Figure \ref{fig:attacked} shows the performance results of the beamforming prediction model's evaluation results under the FGSM attack. We have used $l_{inf}$ norm as the distance metric, which shows the maximum allowable perturbation amount for each item in the input vector $\mathbf{x}$. 

\begin{figure}
    \centering
    \includegraphics[width=0.7\linewidth]{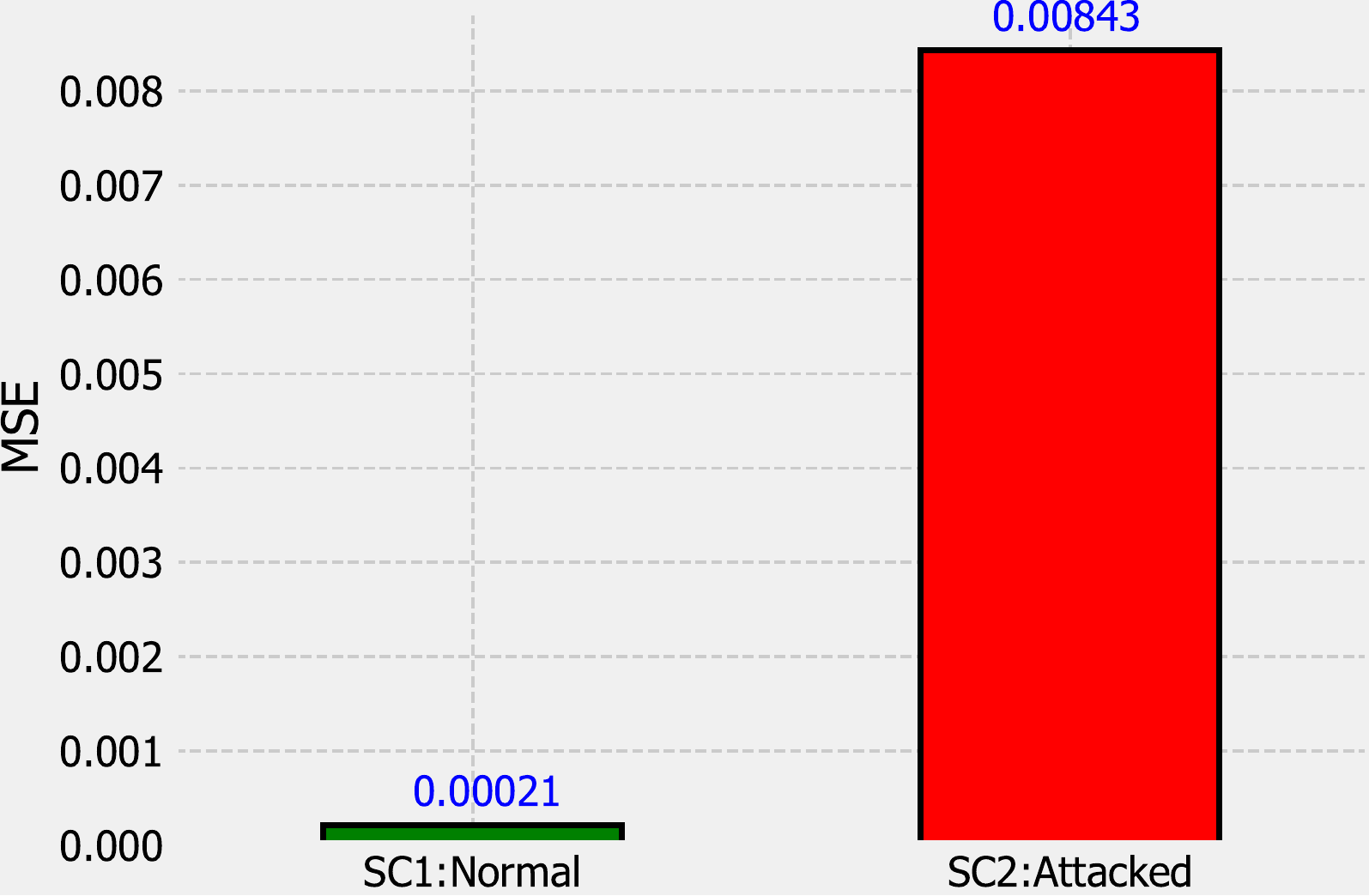}
    \caption{The performance results for SC1 and SC2.}
    \label{fig:rq1}
\end{figure}

Figure \ref{fig:rq1} shows the mean squared error (MSE) of the performance results with normal and attacked beamforming input.

According to the figures, the undefended RF beamforming codeword prediction model is vulnerable for the FGSM attack. The MSE performance result of the model under attack is approximately 40 (i.e. $\frac{0.00843 (Normal)}{0.00021 (Attacked)} \approx 40.14$) times higher.
\subsection{Results for RQ2}\label{sec:res_rq2}
Adversarial training is a popularly advised defense mechanism that proposes generating adversarial instances using the victim model's loss function, and then re-training the model with the newly generated adversarial instances and their respective outputs. This approach has proved to be effective in protecting deep learning models from adversarial machine learning attacks.

\begin{figure}[h]
    \centering
    \includegraphics[width=0.8\linewidth]{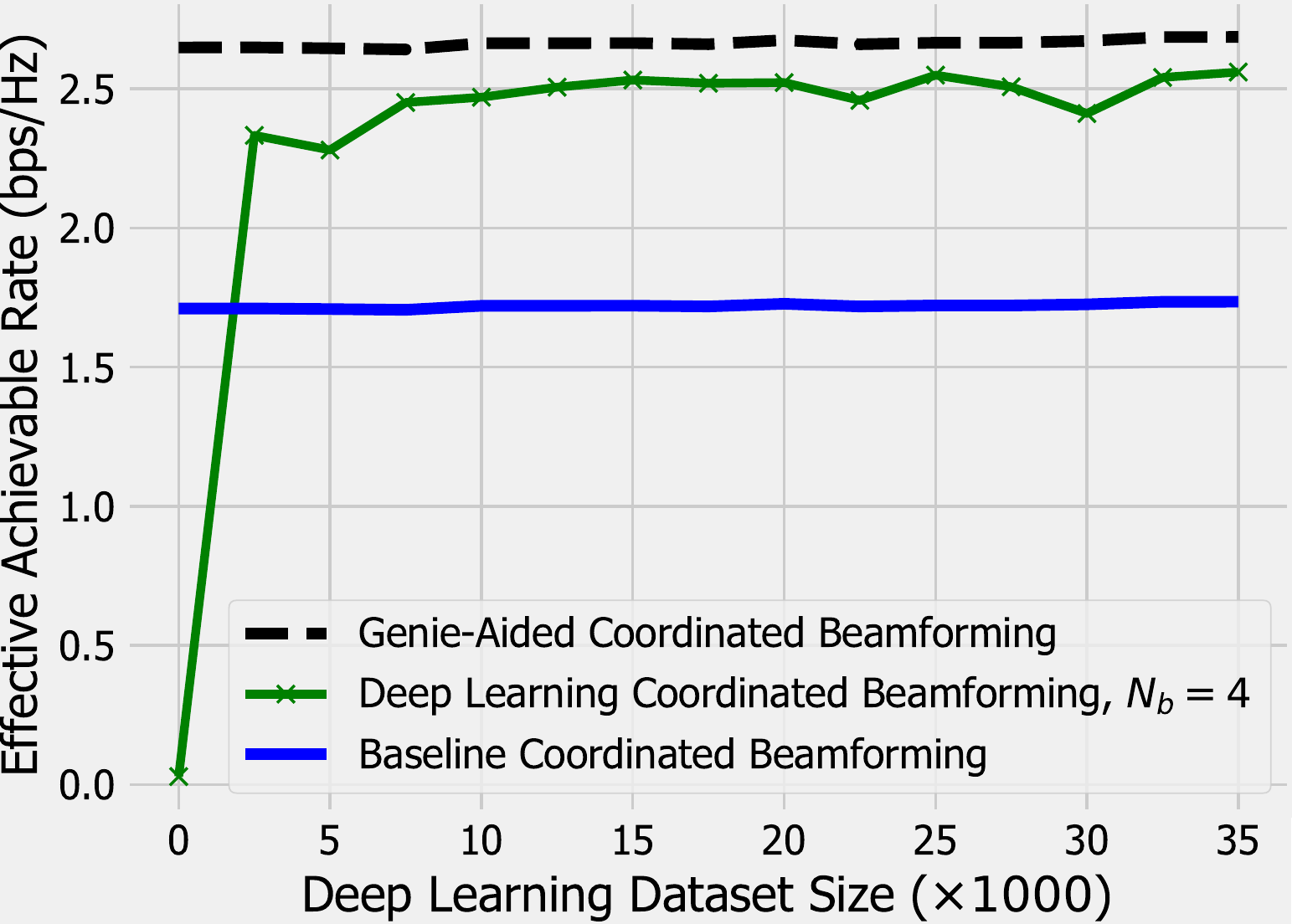}
    \caption{Attacked (Undefended) RF beamforming codeword deep learning model results with adversarial training.}
    \label{fig:adv_training}
\end{figure}

Figure \ref{fig:adv_training} shows the adversarial trained deep learning model results with FGSM attack. According to the figure, the deep learning model's predictions are very close to the original (i.e. undefended and non-attacked) value in Figure \ref{fig:org}.

%\begin{figure}[h]
%    \centering
%    \includegraphics[width=0.8\linewidth]{img/all.pdf}
%    \caption{Attacked (Undefended) RF beamforming codeword deep learning model results with adversarial training.}
%    \label{fig:adv_training}
%\end{figure}

Figure \ref{fig:rq1} shows the MSE of the performance results for all scenarios.
\begin{figure}[h]
    \centering
    \includegraphics[width=0.8\linewidth]{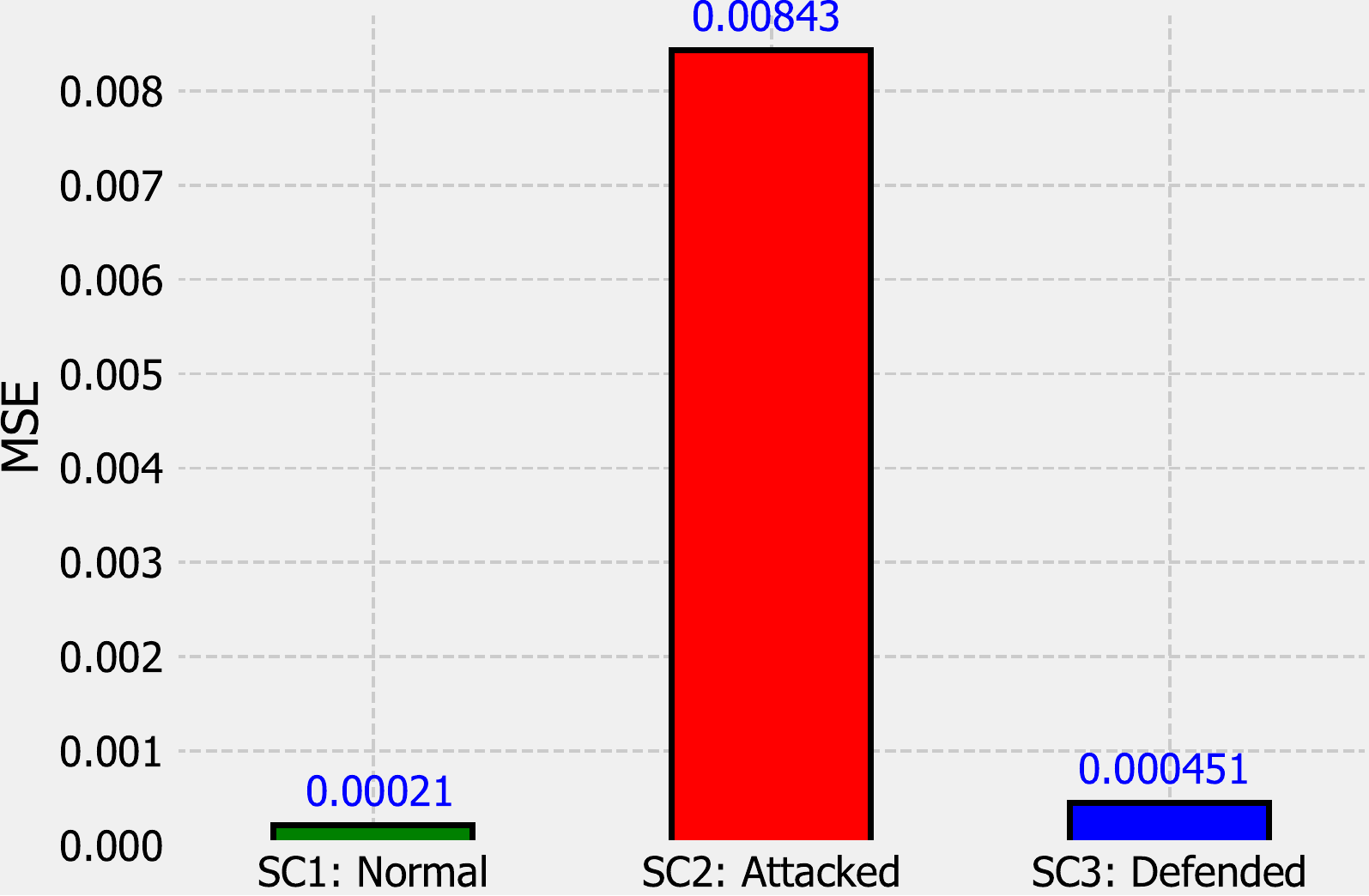}
    \caption{The performance results for all scenarios.}
    \label{fig:my_label}
\end{figure}
\subsection{Threats to Validity}\label{sec:threats_to_val}
A key \textit{external validity} threat is related to the generalization of results \cite{a06430d231c644d8a1278057a2dd6956}. We used only the RF beamforming dataset in our experiments, and we need more case studies to generalize the results. Moreover, the dataset reflects different types of milimeter-wave beams. 

Our key \textit{construct validity} threat is related to the selection of attack type FGSM. Nevertheless, note that this attack is from the literature \cite{a06430d231c644d8a1278057a2dd6956} and applied to several deep learning usage domains. In the future, we will conduct dedicated empirical studies to investigate more adversarial machine learning attacks systematically.

Our main \textit{conclusion validity} threat is due to finding the best attack budget $\epsilon$ that is responsible for manipulating the legitimate user's signal for poisoning the beamforming prediction model. To mitigate this threat, we repeated each experiment 20 times to reduce the probability that the results were obtained by chance. In a standard neural network training, all weights are initialized uniformly at random. In the second stage, using optimization, these weights are updated to fit the classification problem. Since the training started with a probabilistic approach, there is a possibility of facing optimization's local minimum problem. To eliminate the local minimum problem, we repeat the training 20 times to find the $\epsilon$ value that gives the best attack result. In each repetition, the weights were initialized uniformly at random but with different values. If the optimization function failed to find the global minimum in the next experiment, it is likely to see it as the weights have been initialized with different values.
\section{Conclusions and Future Works}\label{sec:conclusion}
This research discussed one of the security issues related to RF beamforming codeword prediction models' vulnerabilities and solutions: (1) Is the deep learning-based RF beamforming codeword predictor vulnerable for adversarial machine learning attacks? (2) Is the iterative adversarial training approach a mitigation method for the adversarial attacks in beamforming prediction? We conducted experiments with the DeepMIMO ray tracing scenario to answer these questions. Our results confirm that the original model is vulnerable to a modified FGSM attack. One of the mitigation methods is the iterative adversarial training approach. Our empirical results also show that iterative adversarial training successfully increases the RF beamforming prediction performance and creates a more accurate predictor, suggesting that the strategy can improve the predictor's performance.  

\bibliographystyle{ieeetr}
\bibliography{myref}

\begin{thebibliography}{10}

\bibitem{9040431}
H.~{Viswanathan} and P.~E. {Mogensen}, ``Communications in the 6g era,'' {\em
  IEEE Access}, vol.~8, pp.~57063--57074, 2020.

\bibitem{9321326}
J.~{Kaur}, M.~A. {Khan}, M.~{Iftikhar}, M.~{Imran}, and Q.~{Emad Ul Haq},
  ``Machine learning techniques for 5g and beyond,'' {\em IEEE Access}, vol.~9,
  pp.~23472--23488, 2021.

\bibitem{7792374}
C.~{Jiang}, H.~{Zhang}, Y.~{Ren}, Z.~{Han}, K.~{Chen}, and L.~{Hanzo},
  ``Machine learning paradigms for next-generation wireless networks,'' {\em
  IEEE Wireless Communications}, vol.~24, no.~2, pp.~98--105, 2017.

\bibitem{8666641}
C.~{Zhang}, P.~{Patras}, and H.~{Haddadi}, ``Deep learning in mobile and
  wireless networking: A survey,'' {\em IEEE Communications Surveys Tutorials},
  vol.~21, no.~3, pp.~2224--2287, 2019.

\bibitem{8972389}
K.~{Shafique}, B.~A. {Khawaja}, F.~{Sabir}, S.~{Qazi}, and M.~{Mustaqim},
  ``Internet of things (iot) for next-generation smart systems: A review of
  current challenges, future trends and prospects for emerging 5g-iot
  scenarios,'' {\em IEEE Access}, vol.~8, pp.~23022--23040, 2020.

\bibitem{8360430}
M.~G. {Kibria}, K.~{Nguyen}, G.~P. {Villardi}, O.~{Zhao}, K.~{Ishizu}, and
  F.~{Kojima}, ``Big data analytics, machine learning, and artificial
  intelligence in next-generation wireless networks,'' {\em IEEE Access},
  vol.~6, pp.~32328--32338, 2018.

\bibitem{6736750}
W.~{Roh}, J.~{Seol}, J.~{Park}, B.~{Lee}, J.~{Lee}, Y.~{Kim}, J.~{Cho},
  K.~{Cheun}, and F.~{Aryanfar}, ``Millimeter-wave beamforming as an enabling
  technology for 5g cellular communications: theoretical feasibility and
  prototype results,'' {\em IEEE Communications Magazine}, vol.~52, no.~2,
  pp.~106--113, 2014.

\bibitem{8395149}
A.~{Alkhateeb}, S.~{Alex}, P.~{Varkey}, Y.~{Li}, Q.~{Qu}, and D.~{Tujkovic},
  ``Deep learning coordinated beamforming for highly-mobile millimeter wave
  systems,'' {\em IEEE Access}, vol.~6, pp.~37328--37348, 2018.

\bibitem{7400949}
R.~W. {Heath}, N.~{González-Prelcic}, S.~{Rangan}, W.~{Roh}, and A.~M.
  {Sayeed}, ``An overview of signal processing techniques for millimeter wave
  mimo systems,'' {\em IEEE Journal of Selected Topics in Signal Processing},
  vol.~10, no.~3, pp.~436--453, 2016.

\bibitem{8767936}
J.~{Mo}, B.~L. {Ng}, S.~{Chang}, P.~{Huang}, M.~N. {Kulkarni}, A.~{Alammouri},
  J.~C. {Zhang}, J.~{Lee}, and W.~J. {Choi}, ``Beam codebook design for 5g
  mmwave terminals,'' {\em IEEE Access}, vol.~7, pp.~98387--98404, 2019.

\bibitem{9048614}
S.~{Chen}, S.~{Sun}, G.~{Xu}, X.~{Su}, and Y.~{Cai}, ``Beam-space multiplexing:
  Practice, theory, and trends, from 4g td-lte, 5g, to 6g and beyond,'' {\em
  IEEE Wireless Communications}, vol.~27, no.~2, pp.~162--172, 2020.

\bibitem{9034044}
M.~S. {Sim}, Y.~{Lim}, S.~H. {Park}, L.~{Dai}, and C.~{Chae}, ``Deep
  learning-based mmwave beam selection for 5g nr/6g with sub-6 ghz channel
  information: Algorithms and prototype validation,'' {\em IEEE Access},
  vol.~8, pp.~51634--51646, 2020.

\bibitem{8644288}
Y.~{Wang}, A.~{Klautau}, M.~{Ribero}, M.~{Narasimha}, and R.~W. {Heath},
  ``Mmwave vehicular beam training with situational awareness by machine
  learning,'' in {\em 2018 IEEE Globecom Workshops (GC Wkshps)}, pp.~1--6,
  2018.

\bibitem{9229146}
J.~{Suomalainen}, A.~{Juhola}, S.~{Shahabuddin}, A.~{Mämmelä}, and
  I.~{Ahmad}, ``Machine learning threatens 5g security,'' {\em IEEE Access},
  vol.~8, pp.~190822--190842, 2020.

\bibitem{sagduyu2021adversarial}
Y.~E. Sagduyu, T.~Erpek, and Y.~Shi, ``Adversarial machine learning for 5g
  communications security,'' {\em arXiv preprint arXiv:2101.02656}, 2021.

\bibitem{7742901}
V.~{Va}, J.~{Choi}, and R.~W. {Heath}, ``The impact of beamwidth on temporal
  channel variation in vehicular channels and its implications,'' {\em IEEE
  Transactions on Vehicular Technology}, vol.~66, no.~6, pp.~5014--5029, 2017.

\bibitem{2016arXiv161101236K}
A.~{Kurakin}, I.~{Goodfellow}, and S.~{Bengio}, ``{Adversarial Machine Learning
  at Scale},'' {\em arXiv e-prints}, p.~arXiv:1611.01236, Nov. 2016.

\bibitem{8965459}
M.~{Aladag}, F.~O. {Catak}, and E.~{Gul}, ``Preventing data poisoning attacks
  by using generative models,'' in {\em 2019 1st International Informatics and
  Software Engineering Conference (UBMYK)}, pp.~1--5, 2019.

\bibitem{2021arXiv210204150F}
O.~{Faruk Tuna}, F.~{Ozgur Catak}, and M.~{Taner Eskil}, ``{Exploiting
  epistemic uncertainty of the deep learning models to generate adversarial
  samples},'' {\em arXiv e-prints}, p.~arXiv:2102.04150, Feb. 2021.

\bibitem{alkhateeb2019deepmimo}
A.~Alkhateeb, ``Deepmimo: A generic deep learning dataset for millimeter wave
  and massive mimo applications,'' 2019.

\bibitem{a06430d231c644d8a1278057a2dd6956}
P.~Runeson, M.~H{\"o}st, R.~Austen, and B.~Regnell, {\em Case Study Research in
  Software Engineering – Guidelines and Examples}.
\newblock United States: John Wiley and Sons Inc., 2012.

\end{thebibliography}

\end{document}